%% file: main.tex
\documentclass[sigconf]{acmart}

\usepackage{subcaption}
\usepackage{booktabs}
\usepackage{graphicx}
\usepackage{multirow}
\usepackage{booktabs}
\usepackage{siunitx} 
\usepackage{booktabs}    
\usepackage{multirow}    
\usepackage{colortbl}    
\usepackage{xcolor}      
\usepackage{array}   
\usepackage{enumitem}

\AtBeginDocument{%
  }

\copyrightyear{2025}
\acmYear{2025}
\setcopyright{cc}
\setcctype{by}
\acmConference[CIKM '25]{Proceedings of the 34th ACM International Conference on Information and Knowledge Management}{November 10--14, 2025}{Seoul, Republic of Korea}
\acmBooktitle{Proceedings of the 34th ACM International Conference on Information and Knowledge Management (CIKM '25), November 10--14, 2025, Seoul, Republic of Korea}\acmDOI{10.1145/3746252.3761422}
\acmISBN{979-8-4007-2040-6/2025/11}

\begin{document}

\title{GSTBench: A Benchmark Study on the Transferability of Graph Self-Supervised Learning}

\author{Yu Song}
\affiliation{%
\department{Computer Science and Engineering}
  \institution{Michigan State University}
  \city{East Lansing}
  \state{MI}
  \country{USA}}
\email{songyu5@msu.edu}

\author{Zhigang Hua}
\affiliation{%
\department{Monetization AI}
  \institution{Meta}
  \city{Sunnyvale}
  \state{CA}
  \country{USA}}
\email{zhua@meta.com}

\author{Yan Xie}
\affiliation{%
\department{Monetization AI}
  \institution{Meta}
  \city{Sunnyvale}
  \state{CA}
  \country{USA}}
\email{yanxie@meta.com}

\author{Jingzhe Liu}
\affiliation{%
 \department{Computer Science and Engineering}
  \institution{Michigan State University}
  \city{East Lansing}
  \state{MI}
  \country{USA}}
\email{liujin33@msu.edu}

\author{Bo Long}
\affiliation{%
  \department{Monetization AI}
  \institution{Meta}
  \city{Menlo Park}
  \state{CA}
  \country{USA}}
\email{bolong@meta.com}

\author{Hui Liu}
\affiliation{%
  \department{Computer Science and Engineering}
  \institution{Michigan State University}
  \city{East Lansing}
  \state{MI}
  \country{USA}}
\email{liuhui7@msu.edu}

\renewcommand{\shortauthors}{Yu Song et al.}

\begin{abstract}
\input{sections/abstract}

\end{abstract}



\maketitle

\section{Introduction}

\input{sections/intro}

\section{Related Work}

\input{sections/related_work}

\section{Preliminaries}

\input{sections/preliminary}

\section{Empirical Studies}
\input{sections/experiments}

\section{Conclusion}
\input{sections/conclusion}

\section*{Acknowledgements}
Yu Song, Jingzhe Liu, and Hui Liu are supported by the National Science Foundation (NSF) under grant numbers CNS2321416, IIS2212032, IIS2212144, IOS2107215, 
DUE2234015, CNS2246050, DRL2405483 and IOS2035472, US Department of Commerce, Gates Foundation, the Michigan Department of Agriculture and Rural Development, Amazon, Meta and SNAP.

\appendix
\input{sections/appendix}

\section*{GenAI Usage Disclosure}

We used generative AI tools (e.g., ChatGPT) solely for the purpose of refining the language and improving the clarity of writing. No content, code, analysis, or experimental results were generated by AI. All technical contributions, empirical findings, and scientific insights are original and authored by the researchers.

\bibliographystyle{ACM-Reference-Format}
\balance
\bibliography{ref}

\end{document}

%% file: sections/abstract.tex
Self-supervised learning (SSL) has shown great promise in graph representation learning. However, most existing graph SSL methods are developed and evaluated under a single-dataset setting, leaving their cross-dataset transferability largely unexplored and limiting their ability to leverage knowledge transfer and large-scale pretraining, factors that are critical for developing generalized intelligence beyond fitting training data.
To address this gap and advance foundation model research for graphs, we present \textbf{GSTBench}, the first systematic benchmark for evaluating the transferability of graph SSL methods. We conduct large-scale pretraining on \texttt{ogbn-papers100M} and evaluate five representative SSL methods across a diverse set of target graphs. Our standardized experimental setup decouples confounding factors such as model architecture, dataset characteristics, and adaptation protocols, enabling rigorous comparisons focused solely on pretraining objectives. 
Surprisingly, we observe that most graph SSL methods struggle to generalize, with some performing worse than random initialization. In contrast, GraphMAE, a masked autoencoder approach, consistently improves transfer performance. We analyze the underlying factors that drive these differences and offer insights to guide future research on transferable graph SSL, laying a solid foundation for the “pretrain-then-transfer” paradigm in graph learning. Our code is available at \url{https://github.com/SongYYYY/GSTBench}.

%% file: sections/intro.tex
In recent years, Self-supervised learning (SSL) has emerged as a powerful paradigm for extracting meaningful representations from unlabeled data by discovering inherent patterns and correlations. Given its effectiveness in leveraging large-scale data without human supervision, SSL has become a cornerstone for the success of foundation models across various domains \cite{devlin2018bert, He2021MaskedAA, Brown2020LanguageMA, CLIP, ma2024neural}.

In graph machine learning, numerous self-supervised learning methods have been developed to train Graph Neural Networks (GNNs)\cite{hou2022graphmae, kipf2016variational, velivckovic2018deep, zhu2020deep, thakoor2021large, ma2021deep}. While these methods demonstrate success in the traditional \emph{one model, one dataset} setting, where models are pretrained and evaluated on the \textit{same} graph, their transferability across \textit{different} datasets remains largely unexplored. Specifically, it is still unclear how effectively current graph SSL methods can generalize when pretrained on one or more source datasets and subsequently applied to novel graph structures and domains.

This limitation represents a significant departure from established practices in computer vision (CV) and natural language processing (NLP), where SSL methods enable training of powerful backbone models on web-scale data to learn generalizable representations that transfer effectively across diverse downstream tasks. The limited understanding of transferability in current graph SSL methods fundamentally constrains their potential for cross-dataset knowledge transfer and large-scale pretraining---capabilities that are pivotal to the development of graph foundation models (GFMs)~\cite{mao2024position, song2025scalable}.

A fundamental barrier to achieving transferable graph SSL lies in the problem of \emph{feature heterogeneity}—the fact that different graph datasets often exhibit distinct node attributes, dimensionalities, and semantic meanings.
Recent work has attempted to overcome this challenge by leveraging large language models (LLMs) to encode node features into a shared semantic space~\cite{chen2024text, song2024pure, liu2023one, huang2024prodigy, wang2024graph}. By transforming textual node attributes into rich, high-dimensional representations, LLMs offer a promising solution for harmonizing input spaces across diverse graphs, potentially enabling large-scale GNN pretraining that mirrors the success of foundation models in CV and NLP.

Despite these advances, a critical gap remains: existing prototypes for graph foundation models often adopt vastly different pipelines, each tailored to a specific combination of design choices~\cite{song2025scalable, sun2023all, zhao2024all, huang2024prodigy, song2024pure, li2024zerog}. As a result, they serve more as isolated case studies than as general, principled frameworks, in contrast to widely adopted paradigms such as next-token prediction in NLP~\cite{Brown2020LanguageMA} or masked image modeling in computer vision~\cite{He2021MaskedAA}. Consequently, follow-up work frequently resorts to ad hoc experimentation with various pretraining methods, resulting in redundant efforts and a lack of systematic guidance.
We attribute this gap to three key challenges:

\begin{itemize}
    \item \textbf{Entangled design pipelines:} Most works on graph foundation models introduce highly intricate frameworks that integrate multiple elements, including pretraining objectives, downstream adaptation modules, task unification strategies, and specialized model architectures~\cite{liu2023one, li2024zerog, huang2024prodigy, xia2024anygraph}. This tight coupling of components makes it difficult to isolate and rigorously assess the contribution of individual design elements, particularly the pretraining task, which is the central focus of our study.

    \item \textbf{Inconsistent evaluation landscape:} The lack of standardized experimental protocols and clearly defined evaluation settings across existing studies prevents fair and meaningful comparisons. As a result, reported gains may stem from varying factors such as dataset-specific tuning or evaluation differences, rather than the quality of the pretraining strategy itself.

    \item \textbf{Insufficient pretraining scale:} Existing studies conduct pretraining on datasets of varying sizes, with most limited to small- or medium-scale graphs. This inconsistency hinders meaningful comparison across methods and fails to reveal the true potential of large-scale pretraining. As a result, conclusions drawn from such experiments often lack generalizability and scalability.

\end{itemize}

To address these challenges, we present \textbf{GSTBench} (\underline{G}raph \underline{S}SL \underline{T}ransferability Benchmark), the first systematic study of graph SSL transferability in LLM-unified feature spaces. Our work establishes a controlled experimental environment that standardizes model architectures, datasets, and evaluation protocols, enabling us to focus solely on the impact of pretraining objectives. Leveraging the \texttt{ogbn-papers100M} dataset~\cite{hu2020openOGB} which comprises more than 100 million nodes as our pretraining corpus, we conduct pretraining experiments at an unprecedented scale. Our comprehensive evaluations across multiple domains and tasks reveal several key findings. Notably, most SSL approaches exhibit unstable effects on unseen datasets, with performance occasionally deteriorating below random initialization. Among those, a masked autoencoding approache, GraphMAE \cite{hou2022graphmae}, consistently demonstrates robust transfer capabilities across datasets and tasks, suggesting that reconstructing the rich LLM features provides a stable pretraining signal. Conversely, contrastive learning methods~\cite{zhu2020deep, velivckovic2018deep}, which heavily rely on data augmentation, often lead to negative transfer even under similar data distributions. In addition, we reveal several critical observations on task transferability, model architecture, domain similarity and adaptation methods. Our key contributions can be summarized as follows:

\begin{itemize}
    \item We introduce \textbf{GSTBench}, the first systematic evaluation of graph SSL transferability in LLM-unified feature spaces, under a controlled and reproducible experimental setup.
    \item We leverage the large-scale \texttt{ogbn-papers100M} dataset as a pretraining corpus to study the behavior of graph SSL methods beyond traditional pretraining scales.
    \item We provide novel empirical insights and practical guidelines for designing more effective and transferable graph pretraining strategies.
\end{itemize}

%% file: sections/related_work.tex
\noindent
\textbf{Graph self-supervised learning (GSSL).}
Self-supervised learning on graphs has emerged as a promising approach to address label scarcity in real-world scenarios. These approaches primarily fall into two broad categories: \textit{contrastive learning} and \textit{generative modeling}. Contrastive methods aim to learn node representations by distinguishing positive node pairs from negative ones via random data augmentations~\cite{velivckovic2018deep, zhu2020deep, you2020graph}. In contrast, generative methods aim to reconstruct node features or graph structures from corrupted inputs, providing a denoising signal for representation learning~\cite{hou2022graphmae, hou2023graphmae2, kipf2016variational, song2024pure}. While these methods have shown success in single-dataset settings, their transferability across different datasets and domains remains largely unexplored, particularly when scaling to large pretraining corpora.

\noindent
\textbf{Graph foundation models (GFMs).}
Inspired by the success of foundation models in NLP and CV, recent research has focused on developing graph foundation models that generalize across multiple datasets and tasks~\cite{liu2023one, song2025scalable, huang2024prodigy, li2024zerog, chen2024llaga, song2024pure, xia2024anygraph}. These approaches often involve addressing  heterogeneity across datasets and tasks, designing effective pretraining objectives, and developing
robust adaptation mechanisms. While some works propose customized pretext tasks, they can largely be viewed as specialized variants of existing graph self-supervised learning methods, where modifications are tailored to accommodate particular model architectures or downstream task requirements. In contrast, our work conducts a systematic and general evaluation of representative SSL methods, independent of specific model designs or downstream settings, to assess their raw transferability and and deliver generalizable guidance. 

\noindent
\textbf{Learning on text-attributed graphs.}
The fusion of textual information with graph-structured data has gained significant traction, driven by the abundance of textual content and the impressive capabilities of large language models~\cite{chien2021node, zhao2022learning, he2023harnessing, li2023grenade, chen2024exploring}. Recent approaches have proposed joint learning frameworks where either LLMs are fine-tuned to incorporate structural signals from graphs, or GNNs are enhanced using feedback derived from LLM-generated embeddings. These methods typically emphasize deep modality interaction and the mutual adaptation of text and structure to improve downstream performance. In contrast, our work decouples this interaction by \textit{freezing} the LLM-derived node features and focusing exclusively on the impact of different graph self-supervised objectives, enabling a principled investigation into the transferability and generalization properties of GSSL methods across heterogeneous graphs with unified textual features.

%% file: sections/preliminary.tex
\vspace{1mm}
\noindent
\textbf{Graph learning with LLM-derived features.}
To address feature heterogeneity and unify input spaces across diverse graphs, recent efforts have leveraged the representational power of large language models (LLMs) to encode raw node attributes into fixed-dimensional embeddings. Formally, let $\mathcal{G} = (\mathcal{V}, \mathcal{E})$ be a graph where each node $v_i \in \mathcal{V}$ is associated with a textual attribute $t_i$. A pretrained LLM $\mathcal{L}$ is used to convert $t_i$ into a dense feature vector:
\[
\mathbf{x}_i = \mathcal{L}(t_i) \in \mathbb{R}^d,
\]
where $d$ is the LLM embedding dimension. This transformation enriches node representations with semantic knowledge, often outperforming traditional shallow features such as bag-of-words or word2vec~\cite{chen2024exploring, mikolov2013efficient}. Moreover, this paradigm can be extended to domains where node attributes are not natively textual but can be described in natural language~\cite{liu2023one}, greatly broadening its applicability across domains and tasks.

\vspace{1mm}
\noindent
\textbf{Graph transfer learning.}
Traditional graph machine learning typically focuses on a single dataset and task. In contrast, \emph{graph transfer learning} aims to pretrain a model on a source graph $\mathcal{G}^s$ and adapt it to a target graph $\mathcal{G}^t$, which may differ in both data distribution and task objective.

Let $f(\cdot;\theta)$ denote a graph neural network (GNN) parameterized by $\theta$, which maps an input graph to latent representations. During pretraining, the model is optimized on $\mathcal{G}^s$ using a self-supervised \emph{pretext objective} $\mathcal{L}^s$. This is typically implemented using a lightweight task-specific head $h(\cdot;\phi)$, leading to the following training objective:
\[
\min_{\theta,\phi} \,\mathcal{L}^s\bigl(f(\cdot;\theta),\, h(\cdot;\phi),\, \mathcal{G}^s\bigr).
\]
After pretraining, the head $h$ is discarded, and only the encoder $f(\cdot;\theta)$ is retained for downstream use. The pretrained model is then adapted to the target task on $\mathcal{G}^t$ via a loss $\mathcal{L}^t$ under an adaptation strategy $\mathcal{A}$, yielding task-specific parameters $\theta^*$.

\vspace{1mm}
\noindent
\textbf{Adaptation strategies.}
Recent studies extend beyond traditional adaptation approaches such as fine-tuning and explore more advanced methods, such as adapter-tuning and in-context learning~\cite{zi2024prog, sun2022gppt, sun2023graph}. For broad applicability, we consider three widely used strategies in our study: linear probing, fine-tuning, and in-context learning, which cover the majority of scenarios and offer a balance between computational cost and task performance. 

\paragraph{Linear Probing.}
In linear probing, the pretrained GNN $f(\cdot;\theta)$ is frozen, and a new classification head $h(\cdot;\phi)$ is trained on the target graph $\mathcal{G}^t$. Specifically,
\[
\theta^* \;=\; \Bigl(\theta,\; \arg\min_{\phi}\,\mathcal{L}^t\bigl(f(\cdot;\theta),\,h(\cdot;\phi),\,\mathcal{G}^t\bigr)\Bigr).
\]
Here, only $\phi$ is updated, providing a computationally light way to assess the linear separability of the learned representations.

\paragraph{Fine-tuning.}
Fine-tuning is a canonical adaptation strategy in which the entire model is optimized toward the target task. It starts from the pretrained initialization, potentially locating a region in parameter space that generalizes well, especially given limited downstream data. Formally, both $\theta$ and $\phi$ are optimized in fine-tuning:
\[
\theta^* \;=\; \arg\min_{\theta,\phi}\,\mathcal{L}^t\bigl(f(\cdot;\theta),\,h(\cdot;\phi),\,\mathcal{G}^t\bigr).
\]
Though more computationally expensive, fine-tuning offers greater flexibility by refining both the feature extractor and the task head, potentially yielding better performance.

\paragraph{In-context Learning (ICL)}
Instead of updating model parameters, in-context learning leverages a small set of labeled examples \(\mathcal{D}_{\text{ctx}}\) to guide predictions through a specially designed prompt \(\mathcal{P}\). With ICL, the predictions for nodes in \(\mathcal{G}\) are given by
\[
\hat{Y} = h\bigl(f(\mathcal{P}(\mathcal{G}^t, \mathcal{D}_{\text{ctx}});\, \theta);\, \phi\bigr)
\]
where the prompt \(\mathcal{P}\) aligns the new task with the model’s pretraining objective. This approach provides flexibility in adapting to new tasks without parameter updates, relying on the GNN’s inherent capacity to generalize from its pretraining.

In short, this framework allows us to leverage a shared feature space to pretrain and adapt a single model across different graphs. In following sections, we present a systematic investigation of graph transfer learning’s effectiveness under various pretraining and adaptation strategies.

\begin{table}[ht]
    \centering
    \caption{Summary of METIS partitions of ogbn-papers100M}  
    \label{tab:partitions}
    \resizebox{\columnwidth}{!}{
    \begin{tabular}{c|c|c|c|c}
        \toprule
        \textbf{\#Graphs} & \textbf{Avg. \#Nodes} & \textbf{Avg. \#Edges} & \textbf{\#Node Range} & \textbf{\#Edge Range} \\
        \midrule
        11105 & 10000.90 & 61357.03 & 303 - 45748 & 328 - 122644 \\
        \bottomrule
    \end{tabular}}
\end{table}

\begin{table}[h]
\caption{Summary of the datasets used in our experiments. MMD refers to Maximum Mean Discrepancy.}
\label{tab:datasets_summary}
\resizebox{\linewidth}{!}{
\begin{tabular}{lrrrrrl}
\toprule
\textbf{Name} & \textbf{\#Nodes} & \textbf{\#Edges} & \textbf{\#Classes} & \textbf{MMD} & \textbf{Homophily} & \textbf{Domain} \\
\midrule
Cora     & 2,708  & 10,556  & 7  & 0.07  & 0.81 & Citation \\
Citeseer & 3,186  & 8,450   & 6  & 0.06  & 0.78 & Citation \\
WikiCS   & 11,701 & 431,726 & 10 & 0.10  & 0.65 & Wikipedia \\
DBLP     & 14,376 & 431,326 & 4  & 0.13  & 0.67 & Citation \\
Pubmed   & 19,717 & 88,648  & 3  & 0.15  & 0.80 & Citation \\
\midrule
Amazon Ratings   & 24,492 & 186,100 & 5  & 0.23  & 0.38 & E-commerce \\
Child    & 76,875 & 2,325,044 & 24 & 0.10  & 0.42 & E-commerce \\
Photo    & 48,362 & 873,782  & 12 & 0.13  & 0.75 & E-commerce \\
\bottomrule
\end{tabular}}
\end{table}

%% file: sections/experiments.tex
In this section, we empirically investigate the transferability of graph self-supervised learning methods. Our study is organized into four parts: experimental settings, an initial examination of SSL transferability, and detailed analyses on node classification and link prediction.

\subsection{Experimental Settings}
\label{sec:exp_set}

\noindent
\textbf{Datasets.} We pretrain all SSL methods on the \texttt{ogbn-papers100M} dataset~\cite{hu2020openOGB}, which contains approximately 100 million nodes and 1.6 billion edges. Due to its large scale, the graph cannot be processed in GPU memory, even on high-end devices such as NVIDIA H100.
 Following~\cite{song2024pure}, we preprocess the dataset into roughly 10,000 subgraphs using the METIS algorithm and apply mini-batch training over these partitions. Statistics for \texttt{ogbn-papers100M} are provided in Table~\ref{tab:partitions}.
For downstream evaluation, we consider eight datasets: five in-domain (Cora, Citeseer, WikiCS, DBLP, Pubmed) and three cross-domain (Amazon Ratings, Child, Photo), as processed in~\cite{chen2024text}. The in-domain datasets are largely from the same domain as the pretraining graph (academic citation networks), with the exception of WikiCS, which is based on Wikipedia articles about computer science. We include it as in-domain due to its structural and semantic similarity. The cross-domain datasets are derived from e-commerce platforms and differ substantially from the pretraining graph in both node features and structural characteristics. 
Node features for all datasets are extracted using SentenceBERT~\cite{reimers2019sentence}, due to its strong empirical performance and widespread use in prior work~\cite{chen2024exploring, song2024pure}.
We report detailed dataset statistics in Table~\ref{tab:datasets_summary}, including Maximum Mean Discrepancy (MMD) with the pretraining dataset (to quantify feature distribution shifts) and the homophily ratio (to capture structural differences).

\vspace{1mm}
\noindent
\textbf{Pretraining Objectives.}  
We study five representative pretraining strategies: two generative methods, two contrastive methods, and one task-specific objective (link prediction). These methods cover diverse philosophies in graph representation learning and have been widely adopted in recent work:

\begin{itemize}
    \item \textbf{GraphMAE}~\cite{hou2022graphmae}: A masked autoencoder that reconstructs node features using local neighborhood information, representing a generative approach focused on semantic feature recovery.
    \item \textbf{VGAE}~\cite{kipf2016variational}: A generative method that reconstructs graph structure using a variational formulation, with KL divergence regularization to enforce smoothness in the latent space.
    \item \textbf{DGI}~\cite{velivckovic2018deep}: A contrastive method that maximizes mutual information between local node embeddings and a global summary, using corrupted inputs as negative samples.
    \item \textbf{GRACE}~\cite{zhu2020deep}: A contrastive method that forms positive and negative pairs via graph augmentations (feature and edge drop), promoting invariance to perturbations.
    \item \textbf{Link Prediction (LP)}~\cite{hamilton2017inductive}: A task-specific objective that trains a GNN to predict edge existence, commonly used as a pretext task.
\end{itemize}

\vspace{1mm}
\noindent
\textbf{Evaluation Settings.}  
After pretraining, we evaluate the obtained model checkpoints on two downstream tasks: node classification and link prediction. For node classification, we follow the few-shot protocol from~\cite{song2024pure}, using five training nodes per class, 500 validation nodes, and the remainder for testing.  Each dataset is evaluated over five random splits, and the mean accuracy (ACC) and standard deviation are reported across these splits. To comprehensively assess the transferability of SSL methods, we evaluate three adaptation strategies on downstream datasets: linear probing, in-context learning, and fine-tuning.
Specifically, we adopt the approach from~\cite{song2024pure} for in-context node classification, utilizing a \textit{prompt graph} with labeled nodes as few-shot examples. A detailed introduction to in-context node classification is presented in Appendix ~\ref{sec:icl__}.

For link prediction, the entire edge set is divided as follows: 40\% for training, 10\% for validation, and 50\% for testing. We report the Mean Reciprocal Rank (MRR) as the evaluation metric, which is averaged over three runs with different random seeds. Following standard practice~\cite{li2023evaluating}, we compute edge embeddings using the Hadamard product of node embeddings, which are passed through a 3-layer MLP to predict edge probabilities. Since link prediction is inherently a self-supervised task (where link existence serves as supervision), we adopt fine-tuning as the adaptation strategy to fully leverage the training signals.

\vspace{1mm}
\noindent
\textbf{Training Details.}  
For pretraining, we use a fixed hyperparameter search range across all experiments. The learning rate is tuned between $1\times10^{-5}$ and $1\times10^{-2}$ using the AdamW optimizer, with a cosine learning rate scheduler and a warmup period during the first epoch. All SSL methods are trained on \texttt{ogbn-papers100M} for a maximum of 5 epochs, as we empirically observe that this is sufficient for convergence. During each epoch, we adopt mini-batch training on the subgraphs extracted from the original \texttt{ogbn-papers100M} dataset, which involves over 10,000 gradient updates. The best checkpoint is selected based on the mean node classification performance on downstream datasets. For SSL-specific hyperparameters such as data augmentation rates, we adhere to the settings used in the original papers and repositories. In downstream adaptation, linear probing experiments use a learning rate of $1\times10^{-2}$ and a weight decay of $1\times10^{-4}$, while fine-tuning experiments involve searching for the optimal learning rate (from $1\times10^{-2}$ to $1\times10^{-4}$) and weight decay (from 0 to $1\times10^{-5}$). We adopt standard Graph Convolutional Networks (GCN)~\cite{kipf2016semi} and Graph Attention Networks (GAT)~\cite{velivckovic2017graph} as our backbone architectures, noting that additional normalization layers and residual connections do not yield improvements in the transfer setting.

\begin{figure*}[t]
    \centering
    \includegraphics[width=\textwidth]{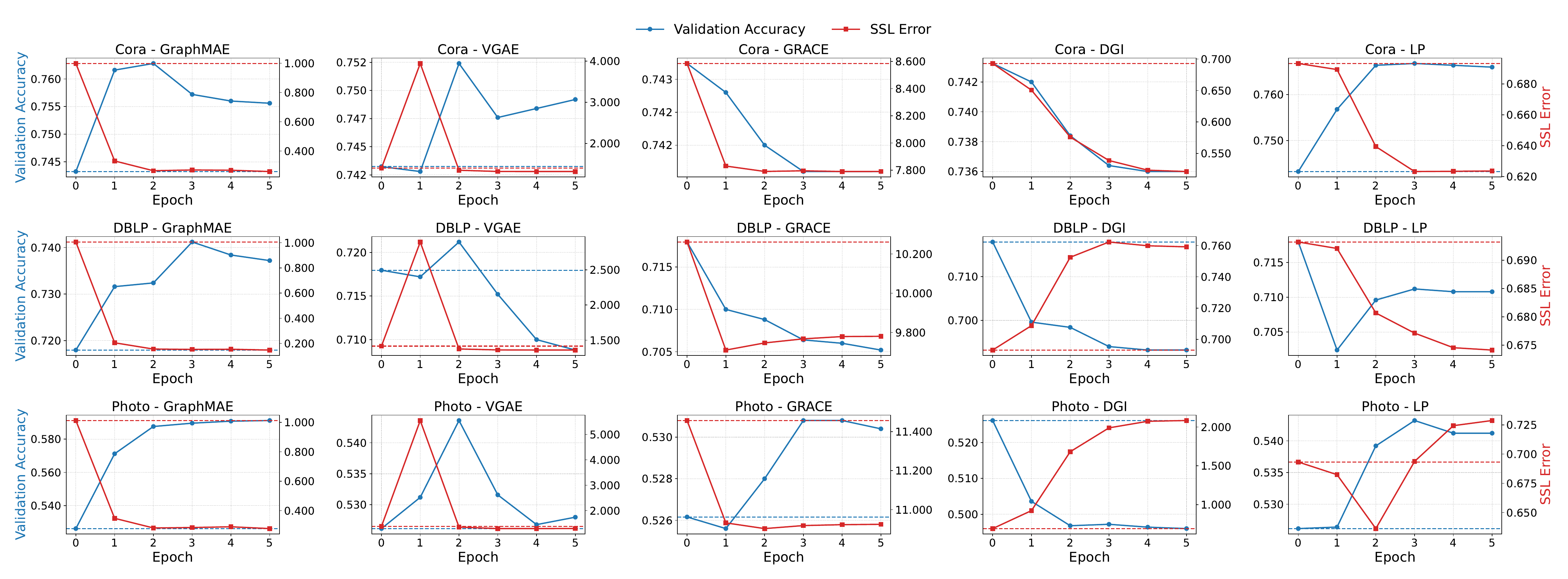} 
    \caption{Different SSL methods exhibit varying behaviors when transferred to unseen datasets. Red line: SSL error on downstream datasets. Blue line: validation accuracy of 5-shot node classification on downstream datasets using linear probing. Epoch 0 indicates a randomly initialized GNN backbone. Among the 5 SSL methods, GraphMAE transfers well to downstream datasets and consistently improves performance, while other methods perform unstably and may cause negative transfer.}
    \label{fig:overview}
\end{figure*}

\subsection{An Overview of SSL Transferability}
In this section, we provide an initial examination of whether pretext tasks transfer to downstream datasets and investigate the correlation between SSL error and downstream performance. 
Figure~\ref{fig:overview} illustrates the progression of SSL error and downstream performance across different pretraining epochs. We show the results for 5 SSL methods, including GraphMAE, VGAE, GRACE, DGI and LP with various datasets. Within each figure, we compare the the SSL error (loss value) evaluated on a given downstream dataset, and the corresponding few-shot classification accuracy. Epoch 0 (and the dashed lines) means the results of a randomly initialized GNN model, referred to as 'baseline'. 
In the following, we draw inspirations from these results via answering a series of research questions and finally provide key takeaways.

\noindent
\textbf{Does the pretext task transfer?} 
We begin by evaluating the transferability of SSL tasks \textit{per se}, i.e., how well a pretrained model can solve its pretext objective on unseen datasets. From the red lines in the plots, it is evident that SSL transferability varies considerably among different methods. For instance, GraphMAE demonstrates a consistent and substantial decrease in SSL loss as pretraining progresses, suggesting that the model's ability to solve the pretext task, i.e., masked feature reconstruction, has been effectively transferred to novel datasets. 
In contrast, other methods exhibit weaker or inconsistent transferability.
 For instance, GRACE exhibits a gradual decrease in SSL loss, but the extent of reduction is much smaller than GraphMAE. VGAE undergoes a surge in error at the beginning of pretraining, while the final loss remains close to that of the untrained baseline, indicating limited generalization of its reconstruction objective. DGI and LP display highly unstable behaviors, with opposite trends observed on different datasets.
 Overall, our observation on SSL errors indicates that pretrained models may not always possess the capability to solve the pretext tasks, depending on the task and datasets.

\vspace{1mm}
\noindent
\textbf{Does pretraining improve downstream performance?} 
Before analyzing the transfer behavior of SSL methods, we first highlight an important observation from the baseline performance at epoch 0 (blue dashed lines in Figure~\ref{fig:overview}). Notably, the classification accuracy achieved by a randomly initialized GNN backbone is already relatively high across several datasets (e.g., $\sim$74\% on Cora). This strong baseline can be largely attributed to the use of LLM-derived node features, which encapsulate rich semantic information learned from extensive pretraining on textual corpora. Consequently, these features are inherently more informative than traditional shallow features such as bag-of-words or handcrafted attributes. Even without pretraining, the GNN acts as a random projection mechanism combined with fixed neighborhood aggregation, which may already be sufficient to extract discriminative patterns from such high-quality inputs. 
Consequently, the performance gain from SSL pretraining, if any, is less about discovering signal from scratch, and more about refining the model's capacity to utilize graph structures.

We then examine how downstream performance evolves throughout pretraining, as shown by the solid blue lines in Figure~\ref{fig:overview}. GraphMAE exhibits consistent improvements over the baseline across multiple datasets, indicating that node feature reconstruction is an effective pretext task for learning transferable and discriminative representations. VGAE, though underperforming GraphMAE, also provides performance gains over the baseline when an appropriate number of pretraining epochs is chosen.
In contrast, the performance of GRACE and LP varies considerably across datasets, with frequent instances of negative transfer. DGI, in particular, consistently leads to performance degradation as pretraining progresses. These observations highlight the importance of carefully selecting pretraining tasks in graph transfer learning. Methods that perform well in their original training context may yield suboptimal or even harmful results when transferred to new graph domains.

\begin{figure}[t]
    \centering
    \begin{subfigure}[b]{0.48\columnwidth}
        \centering
        \includegraphics[width=\linewidth]{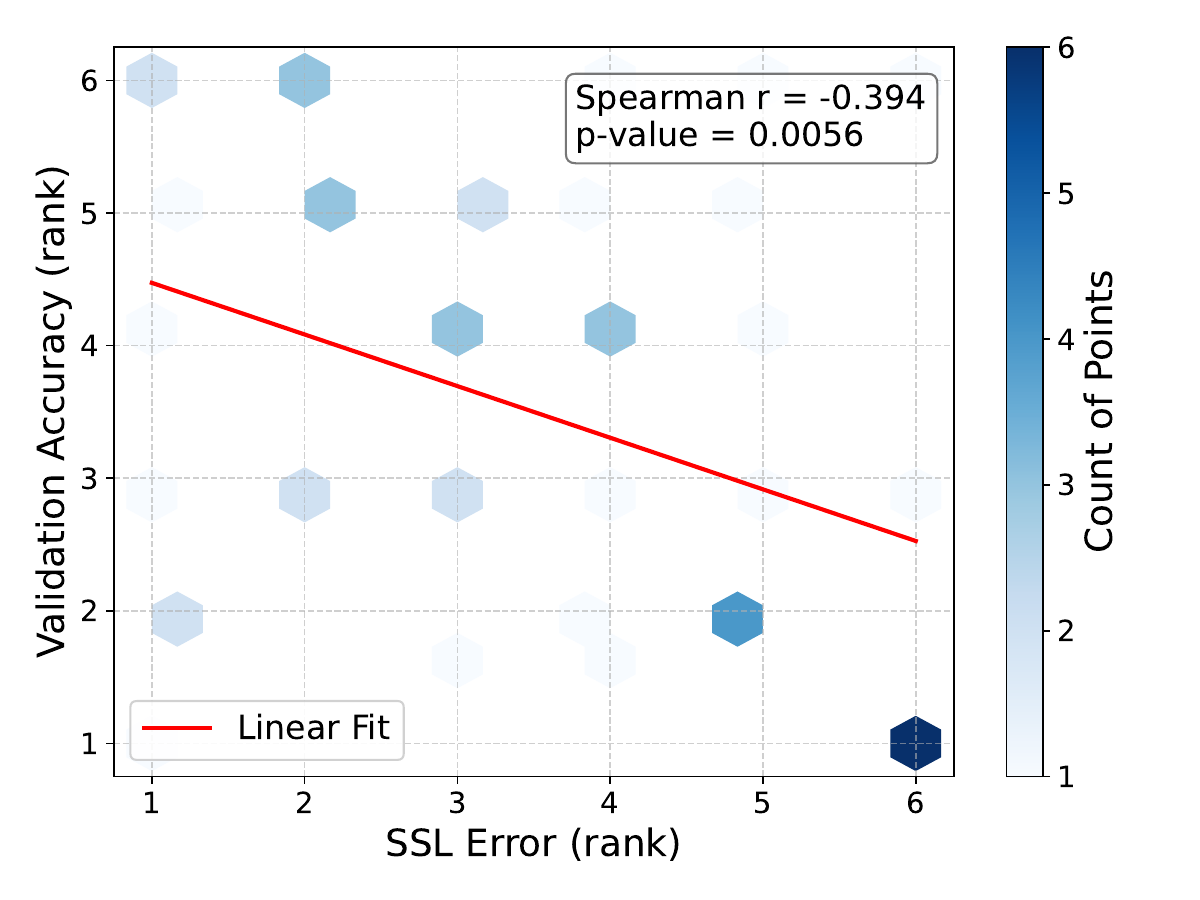} 
        \caption{GCN}
        \label{fig:sub1}
    \end{subfigure}
    \hfill
    \begin{subfigure}[b]{0.48\columnwidth}
        \centering
        \includegraphics[width=\linewidth]{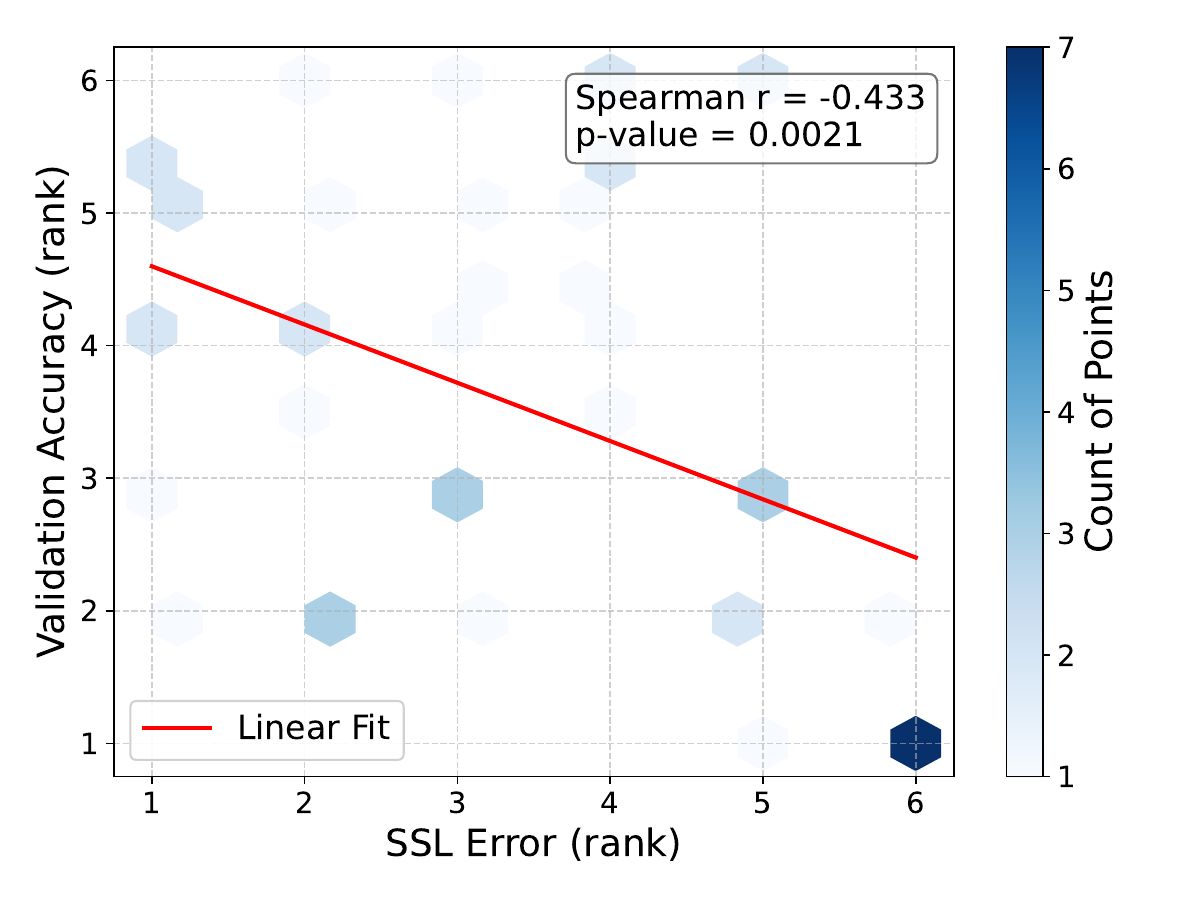} 
        \caption{GAT}
        \label{fig:sub2}
    \end{subfigure}
    \caption{Correlation between SSL error and validation accuracy with GraphMAE. Each hexagon represents a model checkpoint at a specific pretraining epoch, evaluated on a particular dataset. The darkness of the hexagon indicates the density of points at that position. X-axis refers to the rank of SSL error; Y-axis denotes the rank of validation accuracy. The results show that lower SSL error generally leads to better downstream performance, but no strictly.}
    \label{fig:corr}
\end{figure}

\begin{table*}[t]
\caption{Comparison of linear probing accuracy with different pre-training objectives.  Shaded cells mark the better architecture between GCN and GAT.  \textbf{Bold} numbers are the best across GCN for a given dataset.  \underline{Underlined} numbers are the best across GAT for a given dataset.}
\label{tab:linear}
\small
\setlength{\tabcolsep}{4pt}
\resizebox{\textwidth}{!}{
\begin{tabular}{ll ccccc c c c ccc c c}
\toprule
& & \multicolumn{5}{c}{In-domain datasets} & \multicolumn{2}{c}{} & \multicolumn{3}{c}{Cross-domain datasets} & & \\
\cmidrule(r){3-7} \cmidrule(r){10-12}
Task & Arch & Citeseer & Cora & DBLP & PubMed & WikiCS & Mean Acc. & Rank & A-Ratings & Child & Photo & Mean Acc. & Rank \\
\midrule
\multirow{2}{*}{Rand. Init.} 
& GCN & 0.680{\scriptsize$\pm$0.017} 
      & 0.737{\scriptsize$\pm$0.016} 
      & 0.712{\scriptsize$\pm$0.033} 
      & 0.668{\scriptsize$\pm$0.033} 
      & 0.618{\scriptsize$\pm$0.060} 
      & 0.683 & 4.20 
      & \cellcolor{gray!15}0.257{\scriptsize$\pm$0.041} 
      & 0.208{\scriptsize$\pm$0.051} 
      & 0.512{\scriptsize$\pm$0.018} 
      & 0.326 & 4.67 \\
& GAT & \cellcolor{gray!15}0.687{\scriptsize$\pm$0.016} 
      & \cellcolor{gray!15}0.744{\scriptsize$\pm$0.017} 
      & \cellcolor{gray!15}0.731{\scriptsize$\pm$0.028} 
      & \cellcolor{gray!15}0.690{\scriptsize$\pm$0.032} 
      & \cellcolor{gray!15}0.706{\scriptsize$\pm$0.033} 
      & \cellcolor{gray!15}0.711 & 4.00 
      & 0.245{\scriptsize$\pm$0.033} 
      & \cellcolor{gray!15}0.239{\scriptsize$\pm$0.020} 
      & \cellcolor{gray!15}0.569{\scriptsize$\pm$0.023} 
      & \cellcolor{gray!15}0.351 & 4.33 \\

\midrule
\multirow{2}{*}{GraphMAE} 
& GCN & 0.686{\scriptsize$\pm$0.013} 
      & 0.748{\scriptsize$\pm$0.011} 
      & \textbf{0.731}{\scriptsize$\pm$0.031} 
      & \textbf{0.702}{\scriptsize$\pm$0.031} 
      & \textbf{0.704}{\scriptsize$\pm$0.035} 
      & \textbf{0.714} & 1.60 
      & \cellcolor{gray!15}0.256{\scriptsize$\pm$0.036} 
      & \textbf{0.259}{\scriptsize$\pm$0.042} 
      & \textbf{0.575}{\scriptsize$\pm$0.012} 
      & 0.363 & 2.67 \\
& GAT & \cellcolor{gray!15}0.697{\scriptsize$\pm$0.016} 
      & \cellcolor{gray!15}0.762{\scriptsize$\pm$0.015} 
      & \cellcolor{gray!15}\underline{0.739}{\scriptsize$\pm$0.027} 
      & \cellcolor{gray!15}\underline{0.717}{\scriptsize$\pm$0.031} 
      & \cellcolor{gray!15}\underline{0.726}{\scriptsize$\pm$0.032} 
      & \cellcolor{gray!15}\underline{0.728} & 1.40 
      & 0.244{\scriptsize$\pm$0.041} 
      & \cellcolor{gray!15}0.263{\scriptsize$\pm$0.036} 
      & \cellcolor{gray!15}0.588{\scriptsize$\pm$0.020} 
      & \cellcolor{gray!15}0.365 & 3.00 \\

\midrule
\multirow{2}{*}{VGAE} 
& GCN & \cellcolor{gray!15}\textbf{0.697}{\scriptsize$\pm$0.018} 
      & 0.741{\scriptsize$\pm$0.020} 
      & 0.700{\scriptsize$\pm$0.025} 
      & 0.674{\scriptsize$\pm$0.026} 
      & 0.610{\scriptsize$\pm$0.073} 
      & 0.684 & 3.40 
      & \textbf{0.353}{\scriptsize$\pm$0.022} 
      & 0.227{\scriptsize$\pm$0.047} 
      & 0.525{\scriptsize$\pm$0.021} 
      & \textbf{0.368} & 2.00 \\
& GAT & 0.688{\scriptsize$\pm$0.009} 
      & \cellcolor{gray!15}\underline{0.774}{\scriptsize$\pm$0.009} 
      & \cellcolor{gray!15}0.729{\scriptsize$\pm$0.030} 
      & \cellcolor{gray!15}0.691{\scriptsize$\pm$0.036} 
      & \cellcolor{gray!15}0.714{\scriptsize$\pm$0.031} 
      & \cellcolor{gray!15}0.719 & 2.60 
      & \cellcolor{gray!15}\underline{0.354}{\scriptsize$\pm$0.025} 
      & \cellcolor{gray!15}\underline{0.276}{\scriptsize$\pm$0.040} 
      & \cellcolor{gray!15}\underline{0.598}{\scriptsize$\pm$0.035} 
      & \cellcolor{gray!15}\underline{0.409} & 1.00 \\

\midrule
\multirow{2}{*}{GRACE} 
& GCN & \cellcolor{gray!15}0.692{\scriptsize$\pm$0.026} 
      & 0.737{\scriptsize$\pm$0.013} 
      & 0.701{\scriptsize$\pm$0.028} 
      & 0.671{\scriptsize$\pm$0.032} 
      & 0.619{\scriptsize$\pm$0.065} 
      & 0.684 & 3.80 
      & \cellcolor{gray!15}0.262{\scriptsize$\pm$0.051} 
      & 0.203{\scriptsize$\pm$0.045} 
      & 0.514{\scriptsize$\pm$0.012} 
      & 0.326 & 4.67 \\
& GAT & 0.670{\scriptsize$\pm$0.018} 
      & \cellcolor{gray!15}0.744{\scriptsize$\pm$0.023} 
      & \cellcolor{gray!15}0.720{\scriptsize$\pm$0.037} 
      & \cellcolor{gray!15}0.678{\scriptsize$\pm$0.022} 
      & \cellcolor{gray!15}0.708{\scriptsize$\pm$0.036} 
      & \cellcolor{gray!15}0.704 & 4.80 
      & 0.239{\scriptsize$\pm$0.041} 
      & \cellcolor{gray!15}0.240{\scriptsize$\pm$0.026} 
      & \cellcolor{gray!15}0.571{\scriptsize$\pm$0.026} 
      & \cellcolor{gray!15}0.350 & 4.33 \\

\midrule
\multirow{2}{*}{DGI} 
& GCN & 0.663{\scriptsize$\pm$0.024} 
      & 0.737{\scriptsize$\pm$0.017} 
      & 0.692{\scriptsize$\pm$0.046} 
      & 0.672{\scriptsize$\pm$0.020} 
      & 0.601{\scriptsize$\pm$0.053} 
      & 0.673 & 5.20 
      & \cellcolor{gray!15}0.262{\scriptsize$\pm$0.046} 
      & 0.207{\scriptsize$\pm$0.062} 
      & 0.495{\scriptsize$\pm$0.019} 
      & 0.321 & 4.67 \\
& GAT & \cellcolor{gray!15}\underline{0.698}{\scriptsize$\pm$0.015} 
      & \cellcolor{gray!15}0.741{\scriptsize$\pm$0.015} 
      & \cellcolor{gray!15}0.721{\scriptsize$\pm$0.024} 
      & \cellcolor{gray!15}0.703{\scriptsize$\pm$0.038} 
      & \cellcolor{gray!15}0.705{\scriptsize$\pm$0.037} 
      & \cellcolor{gray!15}0.713 & 3.80 
      & 0.246{\scriptsize$\pm$0.053} 
      & \cellcolor{gray!15}0.236{\scriptsize$\pm$0.025} 
      & \cellcolor{gray!15}0.566{\scriptsize$\pm$0.026} 
      & \cellcolor{gray!15}0.349 & 4.67 \\

\midrule
\multirow{2}{*}{LP} 
& GCN & 0.676{\scriptsize$\pm$0.014} 
      & \cellcolor{gray!15}\textbf{0.754}{\scriptsize$\pm$0.016} 
      & 0.700{\scriptsize$\pm$0.033} 
      & \cellcolor{gray!15}0.684{\scriptsize$\pm$0.025} 
      & 0.619{\scriptsize$\pm$0.062} 
      & 0.687 & 2.80 
      & \cellcolor{gray!15}0.274{\scriptsize$\pm$0.041} 
      & 0.211{\scriptsize$\pm$0.025} 
      & 0.526{\scriptsize$\pm$0.024} 
      & 0.337 & 2.33 \\
& GAT & \cellcolor{gray!15}0.697{\scriptsize$\pm$0.013} 
      & 0.722{\scriptsize$\pm$0.022} 
      & \cellcolor{gray!15}0.728{\scriptsize$\pm$0.032} 
      & 0.683{\scriptsize$\pm$0.029} 
      & \cellcolor{gray!15}0.707{\scriptsize$\pm$0.026} 
      & \cellcolor{gray!15}0.708 & 4.40 
      & 0.258{\scriptsize$\pm$0.044} 
      & \cellcolor{gray!15}0.236{\scriptsize$\pm$0.020} 
      & \cellcolor{gray!15}0.581{\scriptsize$\pm$0.023} 
      & \cellcolor{gray!15}0.358 & 3.67 \\

\bottomrule
\end{tabular}
}
\end{table*}

\noindent
\textbf{Does lower SSL error lead to better performance?}
An important question in evaluating SSL methods is whether improvements in the pretext objective translate to higher downstream accuracy. To investigate this, we focus on GraphMAE—the most stable and successful method in our experiments, and analyze the correlation between its SSL loss and validation accuracy.

As shown in Figure~\ref{fig:corr}, we compute the Spearman correlation between SSL error and accuracy across model checkpoints taken at different pretraining epochs and evaluated on various datasets. Each hexagon in the plot represents a checkpoint, with its darkness indicating the density of nearby points. To facilitate consistent comparison across epochs, we convert both SSL error and accuracy to rank values within each pretraining trajectory. The fitted regression line reveals a statistically significant negative correlation ($p < 0.01$), suggesting that, in general, lower SSL error is associated with better downstream performance.
However, this relationship is \textit{not strictly linear}. We observe that accuracy tends to plateau and may even decline after a certain point, despite continued reductions in SSL loss. This pattern indicates that \textit{over-optimizing} the pretext task may lead the model to capture dataset-specific artifacts that do not generalize well to downstream objectives. These findings highlight the need for early stopping or other regularization strategies when applying SSL in transfer settings.

\noindent
\textbf{Key takeaways.}
Our analysis yields two significant findings. \textbf{(1)} GraphMAE is the only method that consistently transfers effectively to novel datasets and improves performance, while other approaches exhibit unstable behavior and frequently lead to negative transfer. \textbf{(2)} Although GraphMAE shows a general negative correlation between SSL error and downstream performance, a lower SSL error does \textit{not} guarantee higher accuracy. Collectively, our results highlight that selecting an appropriate pretext objective and identifying an optimal stopping point during pretraining are crucial strategies for avoiding negative transfer and maximizing the potential of SSL in graph pretraining.

\subsection{Node Classification}
\label{sec:exp_nc}

In this subsection, we investigate how various factors influence downstream node classification (NC) performance. Specifically, we focus on the effects of \emph{(1) pretraining method}, \emph{(2) backbone architecture}, and \emph{(3) adaptation strategy}. We identify general patterns associated with each factor and offer hypotheses to explain the observed behaviors. We begin by reporting results for linear probing and in-context learning (ICL) in Table~\ref{tab:linear} and Table~\ref{tab:icl}, respectively, and then analyze fine-tuning performance in Table~\ref{tab:ft_nc}. Below, we summarize our key findings.

\vspace{1mm}
\noindent
\textbf{How do different SSL methods affect downstream NC?}  
We first examine the linear probing results in Table~\ref{tab:linear}. Our main observation is that GraphMAE consistently leads to positive transfer across all models and datasets, highlighting its robustness and effectiveness in extracting transferable representations. Interestingly, VGAE performs particularly well on cross-domain datasets, achieving substantial gains over the baseline. This may be attributed to the Gaussian prior in its optimization objective, which encourages smoother latent representations, an inductive bias that appears especially advantageous for cross-domain generalization. 

In contrast, other pretraining methods do not show consistent improvements over random initialization. In some cases, they even degrade performance, potentially due to the disruption of semantic information encoded in the LLM-derived features.
Similar trends are observed in the in-context learning results presented in Table~\ref{tab:icl}, with one notable exception: VGAE fails to replicate the improvements seen in the linear probing setting. This discrepancy indicates that training-free adaptation strategies such as ICL may not fully exploit the representations learned during pretraining. In contrast, lightweight adaptation like linear probing can play a crucial role in aligning pretrained embeddings with downstream objectives.

\begin{table*}[t]
\caption{Comparison of in-context learning accuracy with different pre-training objectives.  Shaded cells mark the better architecture between GCN and GAT.  \textbf{Bold} numbers are the best across GCN for a given dataset.  \underline{Underlined} numbers are the best across GAT for a given dataset.}
\label{tab:icl}
\small
\setlength{\tabcolsep}{4pt}
\resizebox{\textwidth}{!}{
\begin{tabular}{ll ccccc c c c ccc c c}
\toprule
& & \multicolumn{5}{c}{In-domain datasets} & \multicolumn{2}{c}{} & \multicolumn{3}{c}{Cross-domain datasets} & & \\
\cmidrule(r){3-7} \cmidrule(r){10-12}
Task & Arch & Citeseer & Cora & DBLP & PubMed & WikiCS & Mean Acc. & Rank & A-Ratings & Child & Photo & Mean Acc. & Rank \\
\midrule
\multirow{2}{*}{Rand. Init.} 
& GCN
& 0.660{\scriptsize$\pm$0.019}
& 0.705{\scriptsize$\pm$0.023}
& 0.683{\scriptsize$\pm$0.045}
& 0.652{\scriptsize$\pm$0.038}
& 0.549{\scriptsize$\pm$0.080}
& 0.650
& 3.00
& 0.212{\scriptsize$\pm$0.030}
& 0.197{\scriptsize$\pm$0.046}
& 0.498{\scriptsize$\pm$0.031}
& 0.302
& 4.33 \\
& GAT 
& \cellcolor{gray!15}0.672{\scriptsize$\pm$0.018}
& \cellcolor{gray!15}0.708{\scriptsize$\pm$0.024}
& \cellcolor{gray!15}0.709{\scriptsize$\pm$0.039}
& \cellcolor{gray!15}0.656{\scriptsize$\pm$0.039}
& \cellcolor{gray!15}0.658{\scriptsize$\pm$0.025}
& \cellcolor{gray!15}0.681
& 3.80
& \cellcolor{gray!15}0.219{\scriptsize$\pm$0.030}
& \cellcolor{gray!15}0.249{\scriptsize$\pm$0.036}
& \cellcolor{gray!15}0.515{\scriptsize$\pm$0.040}
& \cellcolor{gray!15}0.327
& 4.00 \\

\midrule

\multirow{2}{*}{GraphMAE} 
& GCN
& \textbf{0.683}{\scriptsize$\pm$0.020}
& \cellcolor{gray!15}\textbf{0.727}{\scriptsize$\pm$0.023}
& \textbf{0.695}{\scriptsize$\pm$0.051}
& 0.674{\scriptsize$\pm$0.036}
& \textbf{0.587}{\scriptsize$\pm$0.056}
& \textbf{0.673}
& 1.20
& 0.212{\scriptsize$\pm$0.028}
& \textbf{0.232}{\scriptsize$\pm$0.052}
& 0.500{\scriptsize$\pm$0.034}
& \textbf{0.314}
& 3.00 \\
& GAT
& \cellcolor{gray!15}\underline{0.697{\scriptsize$\pm$0.018}}
& \underline{0.721{\scriptsize$\pm$0.023}}
& \cellcolor{gray!15}\underline{0.715{\scriptsize$\pm$0.041}}
& \cellcolor{gray!15}\underline{0.677{\scriptsize$\pm$0.035}}
& \cellcolor{gray!15}0.668{\scriptsize$\pm$0.032}
& \cellcolor{gray!15}\underline{0.696}
& 1.20
& \cellcolor{gray!15}\underline{0.225{\scriptsize$\pm$0.033}}
& \cellcolor{gray!15}\underline{0.268{\scriptsize$\pm$0.036}}
& \cellcolor{gray!15}\underline{0.530{\scriptsize$\pm$0.043}}
& \cellcolor{gray!15}\underline{0.341}
& 1.33 \\

\midrule

\multirow{2}{*}{VGAE} 
& GCN
& 0.656{\scriptsize$\pm$0.034}
& 0.700{\scriptsize$\pm$0.020}
& 0.677{\scriptsize$\pm$0.052}
& 0.645{\scriptsize$\pm$0.032}
& 0.569{\scriptsize$\pm$0.064}
& 0.649
& 4.00
& 0.210{\scriptsize$\pm$0.016}
& 0.208{\scriptsize$\pm$0.052}
& \textbf{0.507}{\scriptsize$\pm$0.022}
& 0.308
& 3.33 \\
& GAT
& \cellcolor{gray!15}0.662{\scriptsize$\pm$0.015}
& \cellcolor{gray!15}0.719{\scriptsize$\pm$0.030}
& \cellcolor{gray!15}0.698{\scriptsize$\pm$0.048}
& \cellcolor{gray!15}0.649{\scriptsize$\pm$0.033}
& \cellcolor{gray!15}\underline{0.669{\scriptsize$\pm$0.024}}
& \cellcolor{gray!15}0.679
& 3.80
& \cellcolor{gray!15}0.212{\scriptsize$\pm$0.040}
& \cellcolor{gray!15}0.252{\scriptsize$\pm$0.036}
& \cellcolor{gray!15}\underline{0.530{\scriptsize$\pm$0.037}}
& \cellcolor{gray!15}0.331
& 2.67 \\

\midrule

\multirow{2}{*}{GRACE} 
& GCN
& 0.655{\scriptsize$\pm$0.030}
& 0.699{\scriptsize$\pm$0.028}
& 0.674{\scriptsize$\pm$0.058}
& \cellcolor{gray!15}0.646{\scriptsize$\pm$0.032}
& 0.573{\scriptsize$\pm$0.073}
& 0.649
& 4.20
& \cellcolor{gray!15}0.213{\scriptsize$\pm$0.015}
& 0.202{\scriptsize$\pm$0.042}
& 0.503{\scriptsize$\pm$0.022}
& 0.306
& 3.00 \\
& GAT
& \cellcolor{gray!15}0.664{\scriptsize$\pm$0.013}
& \cellcolor{gray!15}0.700{\scriptsize$\pm$0.028}
& \cellcolor{gray!15}0.698{\scriptsize$\pm$0.047}
& 0.638{\scriptsize$\pm$0.030}
& \cellcolor{gray!15}0.644{\scriptsize$\pm$0.026}
& \cellcolor{gray!15}0.669
& 5.60
& 0.210{\scriptsize$\pm$0.038}
& \cellcolor{gray!15}0.252{\scriptsize$\pm$0.039}
& \cellcolor{gray!15}0.529{\scriptsize$\pm$0.037}
& \cellcolor{gray!15}0.331
& 3.33 \\

\midrule

\multirow{2}{*}{DGI} 
& GCN
& 0.651{\scriptsize$\pm$0.024}
& \cellcolor{gray!15}0.720{\scriptsize$\pm$0.031}
& 0.644{\scriptsize$\pm$0.067}
& 0.650{\scriptsize$\pm$0.038}
& 0.508{\scriptsize$\pm$0.101}
& 0.635
& 4.80
& \cellcolor{gray!15}\textbf{0.224}{\scriptsize$\pm$0.018}
& 0.175{\scriptsize$\pm$0.021}
& 0.485{\scriptsize$\pm$0.035}
& 0.295
& 4.33 \\
& GAT
& \cellcolor{gray!15}0.677{\scriptsize$\pm$0.023}
& 0.712{\scriptsize$\pm$0.020}
& \cellcolor{gray!15}0.703{\scriptsize$\pm$0.039}
& \cellcolor{gray!15}0.669{\scriptsize$\pm$0.024}
& \cellcolor{gray!15}0.659{\scriptsize$\pm$0.032}
& \cellcolor{gray!15}0.684
& 3.20
& 0.208{\scriptsize$\pm$0.041}
& \cellcolor{gray!15}0.244{\scriptsize$\pm$0.039}
& \cellcolor{gray!15}0.516{\scriptsize$\pm$0.043}
& \cellcolor{gray!15}0.323
& 5.33 \\

\midrule

\multirow{2}{*}{LP} 
& GCN
& 0.653{\scriptsize$\pm$0.022}
& \cellcolor{gray!15}0.704{\scriptsize$\pm$0.033}
& 0.666{\scriptsize$\pm$0.059}
& \cellcolor{gray!15}\textbf{0.675}{\scriptsize$\pm$0.026}
& 0.565{\scriptsize$\pm$0.081}
& 0.652
& 3.80
& \cellcolor{gray!15}0.219{\scriptsize$\pm$0.028}
& 0.208{\scriptsize$\pm$0.052}
& 0.494{\scriptsize$\pm$0.027}
& 0.307
& 3.00 \\
& GAT
& \cellcolor{gray!15}0.689{\scriptsize$\pm$0.010}
& 0.680{\scriptsize$\pm$0.026}
& \cellcolor{gray!15}0.710{\scriptsize$\pm$0.038}
& 0.654{\scriptsize$\pm$0.040}
& \cellcolor{gray!15}0.661{\scriptsize$\pm$0.029}
& \cellcolor{gray!15}0.679
& 3.40
& 0.213{\scriptsize$\pm$0.028}
& \cellcolor{gray!15}0.244{\scriptsize$\pm$0.041}
& \cellcolor{gray!15}0.524{\scriptsize$\pm$0.049}
& \cellcolor{gray!15}0.327
& 4.33 \\
\bottomrule
\end{tabular}}
\end{table*}

\noindent
\textbf{What causes the differences in SSL transferability?}
From Table~\ref{tab:linear} and Table~\ref{tab:icl}, we observe that \emph{generative SSL} approaches (GraphMAE and VGAE) generally outperform other methods in both effectiveness and stability across datasets. 
This is likely due to their generative objective, which enforces a robust understanding of the underlying data distribution, as evidenced by the widespread success of generation-based pretraining such as MIM (Masked Image Modeling)~\cite{He2021MaskedAA} and MLM (Masked Language Modeling)~\cite{devlin2018bert} in CV and NLP. The \emph{masked feature reconstruction} adopted by GraphMAE works particularly well, achieving best overall  performance. We hypothesize that GraphMAE excels because it accomplishes two objectives in a synergistical manner: it compels the model to learn interaction patterns between distinct nodes while preserving the rich semantics inherent in the LLM-produced node features.
In contrast, \emph{contrastive SSL} methods such as DGI and GRACE focus on learning invariances within the data, under the assumption that augmentations (e.g., edge/feature perturbations) do not significantly alter the intrinsic semantics of samples. This learning strategy, however, may suffer from overfitting the pretraining data, due to suboptimal augmentation configurations, inappropriate perturbation strengths, or insufficient exploration of negative samples~\cite{qi2023contrast, he2020momentum}. Consequently, their effectiveness can be less stable in transfer settings.

\noindent
\textbf{Can pretraining improve performance with fine-tuning?} 
We investigate whether traditional fine-tuning benefits from pretraining, particularly when utilizing a large-scale graph as pretraining data. Table~\ref{tab:ft_nc} presents a comparison between GCNs trained from scratch and those fine-tuned from checkpoints obtained through various pretraining objectives, under the same five-shot classification setting. Unlike our previous findings with linear probing and ICL, fine-tuning shows no significant performance differences across all initialization methods, regardless of whether pretraining was employed. These results indicate that the encoder cannot effectively leverage large-scale pretraining for node classification, even with limited supervision from downstream data. This finding aligns with the observation that most existing work in graph pretraining tends to adopt linear probing and in-context learning for evaluation rather than fine-tuning approaches~\cite{liu2023one, huang2024prodigy, song2024pure, li2024zerog}.

\noindent
\textbf{Which adaptation method should I choose?}
Comparing Table~\ref{tab:linear} (linear probing) and Table~\ref{tab:icl} (ICL), we find that linear probing often provides more stable gains for top-performing SSL methods (notably GraphMAE and VGAE), particularly on cross-domain data. This arises from the well-separated representations produced by the pretrained encoder, along with the data-specific mapping learned by the linear classifier from downstream data. While ICL can still be effective, its training-free nature occasionally destabilizes pretrained model in fully utilizing the pretrained knowledge. Fine-tuning, on the other hand, fails to benefit from pretraining.
Ultimately, \emph{linear probing} is simpler and generally more robust, whereas \emph{ICL} is preferable under strict computation limitations.

\noindent
\textbf{Does the backbone model affect performance?}
We compare the two backbones, GCN and GAT, under linear probing and in-context learning (Table~\ref{tab:linear} and Table~\ref{tab:icl}). 
From the tables, GAT often yields higher absolute accuracies, as its attention mechanism can better capture neighbor relevance when aggregating neighboring information. We hypothesize that the attention mechanism may be particularly useful with the LLM embeddings as features, which contain rich semantics and are suitable for attention estimation. 

\begin{table}
\caption{Fine-tuning accuracy of five-shot node classification.}
\label{tab:ft_nc}
\resizebox{\linewidth}{!}{
\begin{tabular}{lcccccc}
\toprule
\textbf{Method} & \textbf{Cora} & \textbf{Citeseer} & \textbf{WikiCS} & \textbf{DBLP} & \textbf{Pubmed} & \textbf{Mean} \\
\midrule
Random & 0.751{\scriptsize±0.011} & 0.708{\scriptsize±0.020} & 0.706{\scriptsize±0.030} & 0.739{\scriptsize±0.030} & 0.707{\scriptsize±0.031} & 0.722 \\
GraphMAE & 0.750{\scriptsize±0.006} & 0.704{\scriptsize±0.011} & 0.712{\scriptsize±0.034} & 0.735{\scriptsize±0.025} & 0.700{\scriptsize±0.038} & 0.720 \\
VGAE & 0.751{\scriptsize±0.018} & 0.706{\scriptsize±0.026} & 0.693{\scriptsize±0.024} & 0.741{\scriptsize±0.029} & 0.707{\scriptsize±0.046} & 0.720 \\
GRACE & 0.754{\scriptsize±0.012} & 0.708{\scriptsize±0.026} & 0.699{\scriptsize±0.026} & 0.739{\scriptsize±0.027} & 0.699{\scriptsize±0.042} & 0.720 \\
DGI & 0.752{\scriptsize±0.011} & 0.703{\scriptsize±0.021} & 0.704{\scriptsize±0.028} & 0.739{\scriptsize±0.032} & 0.703{\scriptsize±0.047} & 0.720 \\
LP & 0.742{\scriptsize±0.014} & 0.703{\scriptsize±0.024} & 0.698{\scriptsize±0.028} & 0.741{\scriptsize±0.024} & 0.707{\scriptsize±0.041} & 0.718 \\
\bottomrule
\end{tabular}}
\end{table}

\vspace{1mm}
\noindent
\textbf{Key takeaways.}
\textbf{(1)} \emph{Generative SSL} (GraphMAE, VGAE) generally outperforms contrastive learning methods. Feature reconstruction (GraphMAE) is especially effective thanks to rich LLM-derived node features.
\textbf{(2)} GraphMAE and VGAE deliver noticeable positive gains in cross-domain transfer, demonstrating the potential of utilizing pretraining as a way to alleviate data scarcity.
\textbf{(3)} \emph{Linear probing} and \emph{in-context learning} are effective ways to leverage pretrained knowledge, with different performance and computation requirements. In contrast, \emph{Fine-tuning} offers no significant benefit.

\begin{table*}[t]
\caption{Comparison of link prediction MRR with different pre-training objectives.  Shaded cells mark the better architecture between GCN and GAT.  \textbf{Bold} numbers are the best across GCN for a given dataset.  \underline{Underlined} numbers are the best across GAT for a given dataset.}
\label{tab:lp}
\small
\setlength{\tabcolsep}{3.5pt}
\resizebox{\textwidth}{!}{
\begin{tabular}{ll ccccc c c c ccc c c}
\toprule
& & \multicolumn{5}{c}{In-domain datasets} & \multicolumn{2}{c}{} & \multicolumn{3}{c}{Cross-domain datasets} & & \\
\cmidrule(r){3-7} \cmidrule(r){10-12}
Method & Arch & Citeseer & Cora & DBLP & PubMed & WikiCS & Mean MRR & Rank & Amazon & Child & Photo & Mean MRR & Rank \\
\midrule
\multirow{2}{*}{Rand. Init.} 
& GCN 
& 0.594{\scriptsize±0.0034} 
& 0.525{\scriptsize±0.0059} 
& 0.966{\scriptsize±0.0004} 
& \cellcolor{gray!15}{0.710}{\scriptsize±0.0018} 
& \cellcolor{gray!15}{0.803}{\scriptsize±0.0017}
& 0.720 
& 3.60 
& \textbf{\cellcolor{gray!15}{0.921}}{\scriptsize±0.0009}
& \cellcolor{gray!15}{0.808}{\scriptsize±0.0004}
& \textbf{\cellcolor{gray!15}{0.846}}{\scriptsize±0.0002}
& \textbf{\cellcolor{gray!15}{0.858}}
& 2.33 \\

& GAT 
& \cellcolor{gray!15}{0.623}{\scriptsize±0.0032} 
& \cellcolor{gray!15}{0.549}{\scriptsize±0.0033} 
& \underline{\cellcolor{gray!15}{0.967}}{\scriptsize±0.0010} 
& \underline{0.677}{\scriptsize±0.0028} 
& 0.793{\scriptsize±0.0015}
& \underline{\cellcolor{gray!15}{0.722}}
& 2.40 
& 0.918{\scriptsize±0.0001}
& 0.806{\scriptsize±0.0001}
& \underline{0.843}{\scriptsize±0.0011}
& \underline{0.856}
& 3.00 \\
\midrule

\multirow{2}{*}{MAE} 
& GCN 
& \textbf{\cellcolor{gray!15}{0.669}}{\scriptsize±0.0023}
& \textbf{\cellcolor{gray!15}{0.590}}{\scriptsize±0.0022}
& \cellcolor{gray!15}{0.966}{\scriptsize±0.0010}
& \textbf{\cellcolor{gray!15}{0.716}}{\scriptsize±0.0002}
& \textbf{\cellcolor{gray!15}{0.805}}{\scriptsize±0.0004}
& \textbf{\cellcolor{gray!15}{0.749}}
& 1.60
& \textbf{\cellcolor{gray!15}{0.921}}{\scriptsize±0.0021}
& \textbf{\cellcolor{gray!15}{0.810}}{\scriptsize±0.0003}
& \cellcolor{gray!15}{0.844}{\scriptsize±0.0005}
& \textbf{\cellcolor{gray!15}{0.858}}
& 2.33 \\

& GAT 
& 0.629{\scriptsize±0.0020}
& \underline{0.555}{\scriptsize±0.0011}
& 0.966{\scriptsize±0.0013}
& 0.673{\scriptsize±0.0022}
& 0.790{\scriptsize±0.0007}
& \underline{0.722}
& 3.60
& 0.918{\scriptsize±0.0017}
& 0.807{\scriptsize±0.0004}
& 0.840{\scriptsize±0.0003}
& 0.855
& 3.67 \\
\midrule

\multirow{2}{*}{VGAE} 
& GCN 
& 0.500{\scriptsize±0.0519}
& 0.476{\scriptsize±0.0020}
& \cellcolor{gray!15}{0.965}{\scriptsize±0.0012}
& \cellcolor{gray!15}{0.708}{\scriptsize±0.0021}
& \cellcolor{gray!15}{0.803}{\scriptsize±0.0009}
& 0.691
& 5.20
& \cellcolor{gray!15}{0.919}{\scriptsize±0.0015}
& \cellcolor{gray!15}{0.808}{\scriptsize±0.0007}
& \cellcolor{gray!15}{0.844}{\scriptsize±0.0007}
& \cellcolor{gray!15}{0.857}
& 5.67 \\

& GAT 
& \cellcolor{gray!15}{0.592}{\scriptsize±0.0114}
& \cellcolor{gray!15}{0.529}{\scriptsize±0.0055}
& 0.965{\scriptsize±0.0002}
& 0.668{\scriptsize±0.0020}
& 0.792{\scriptsize±0.0000}
& \cellcolor{gray!15}{0.709}
& 5.80
& 0.917{\scriptsize±0.0002}
& 0.806{\scriptsize±0.0006}
& 0.842{\scriptsize±0.0014}
& 0.855
& 4.00 \\
\midrule

\multirow{2}{*}{GRACE} 
& GCN 
& 0.598{\scriptsize±0.0043}
& 0.521{\scriptsize±0.0021}
& 0.966{\scriptsize±0.0010}
& \cellcolor{gray!15}{0.708}{\scriptsize±0.0008}
& \cellcolor{gray!15}{0.803}{\scriptsize±0.0009}
& 0.719
& 4.80
& \cellcolor{gray!15}{0.920}{\scriptsize±0.0015}
& \cellcolor{gray!15}{0.807}{\scriptsize±0.0003}
& \textbf{\cellcolor{gray!15}{0.846}}{\scriptsize±0.0006}
& \textbf{\cellcolor{gray!15}{0.858}}
& 3.67 \\

& GAT 
& \cellcolor{gray!15}{0.620}{\scriptsize±0.0054}
& \cellcolor{gray!15}{0.548}{\scriptsize±0.0035}
& \cellcolor{gray!15}{0.966}{\scriptsize±0.0007}
& 0.676{\scriptsize±0.0017}
& \underline{0.794}{\scriptsize±0.0007}
& \cellcolor{gray!15}{0.721}
& 3.00
& \underline{0.918}{\scriptsize±0.0007}
& 0.806{\scriptsize±0.0014}
& 0.842{\scriptsize±0.0003}
& \underline{0.856}
& 3.00 \\
\midrule

\multirow{2}{*}{DGI} 
& GCN 
& 0.595{\scriptsize±0.0048}
& 0.522{\scriptsize±0.0021}
& \textbf{\cellcolor{gray!15}{0.966}}{\scriptsize±0.0011}
& \cellcolor{gray!15}{0.709}{\scriptsize±0.0003}
& \cellcolor{gray!15}{0.803}{\scriptsize±0.0003}
& 0.719
& 3.20
& \cellcolor{gray!15}{0.920}{\scriptsize±0.0005}
& \cellcolor{gray!15}{0.809}{\scriptsize±0.0004}
& \cellcolor{gray!15}{0.845}{\scriptsize±0.0007}
& \textbf{\cellcolor{gray!15}{0.858}}
& 3.00 \\

& GAT 
& \cellcolor{gray!15}{0.626}{\scriptsize±0.0057}
& \cellcolor{gray!15}{0.546}{\scriptsize±0.0043}
& 0.966{\scriptsize±0.0010}
& 0.675{\scriptsize±0.0032}
& 0.793{\scriptsize±0.0025}
& \cellcolor{gray!15}{0.721}
& 3.20
& \underline{0.918}{\scriptsize±0.0023}
& 0.806{\scriptsize±0.0013}
& 0.842{\scriptsize±0.0004}
& 0.855
& 5.00 \\
\midrule

\multirow{2}{*}{LP} 
& GCN 
& \cellcolor{gray!15}{0.643}{\scriptsize±0.0033}
& \cellcolor{gray!15}{0.574}{\scriptsize±0.0028}
& \cellcolor{gray!15}{0.966}{\scriptsize±0.0004}
& \cellcolor{gray!15}{0.709}{\scriptsize±0.0015}
& \textbf{\cellcolor{gray!15}{0.805}}{\scriptsize±0.0005}
& \cellcolor{gray!15}{0.739}
& 2.60
& \cellcolor{gray!15}{0.919}{\scriptsize±0.0009}
& \cellcolor{gray!15}{0.808}{\scriptsize±0.0006}
& \cellcolor{gray!15}{0.845}{\scriptsize±0.0010}
& \cellcolor{gray!15}{0.857}
& 4.00 \\

& GAT 
& \underline{0.631}{\scriptsize±0.0013}
& 0.543{\scriptsize±0.0013}
& 0.966{\scriptsize±0.0015}
& 0.675{\scriptsize±0.0005}
& \underline{0.794}{\scriptsize±0.0017}
& \underline{0.722}
& 3.00
& \underline{0.918}{\scriptsize±0.0009}
& \underline{0.807}{\scriptsize±0.0005}
& 0.842{\scriptsize±0.0010}
& \underline{0.856}
& 2.33 \\
\bottomrule
\end{tabular}
}
\end{table*}

\subsection{Link Prediction}
\label{sec:exp_lp}

In this subsection, we shift our focus to link prediction (LP). To fully leverage the structural patterns in downstream datasets, we adopt fine-tuning as the adaptation strategy. Table~\ref{tab:lp} reports the test MRR of various SSL pretraining methods. As in prior sections, we structure our analysis around several key research questions.

\vspace{1mm}
\noindent
\textbf{How do different SSL pretraining methods affect LP performance?}  
We first observe that GraphMAE again achieves the highest overall MRR on in-domain datasets, mirroring its strong performance in node classification. By reconstructing semantic embeddings, GraphMAE produces node representations that retain rich, context-aware signals, which benefit both node classification and edge prediction tasks.
Following GraphMAE, Link Prediction (LP) pretraining ranks second, also outperforming the baseline by a notable margin. This result highlights the benefit of task alignment, as LP pretraining essentially mirrors the downstream objective.
DGI and GRACE perform modestly, with MRRs close to or slightly above random initialization, suggesting that contrastive methods yield neutral or limited benefit for LP tasks. 
Surprisingly, VGAE performs worst among all methods, indicating severe negative transfer. 

\vspace{1mm}
\noindent
\textbf{What causes the difference between pretraining with VGAE and LP?}  
Although VGAE and LP share similar pretraining objectives (predicting edge existence), their impact on downstream LP differs significantly: LP pretraining improves performance, whereas VGAE leads to degradation.
We hypothesize that VGAE's poor performance stems from the architectural misalignment between the pretraining and fine-tuning stages. Specifically, VGAE uses a nonparametric \emph{dot-product} decoder during pretraining, while LP employs a learnable \emph{MLP decoder}, which is consistent with the architecture used during fine-tuning. This discrepancy likely introduces different inductive biases into the GNN encoder, resulting in learned representations that are not well-suited for one another.
This finding reveals an implicit pitfall that transferability depends not only on the choice of pretraining objective but also on architectural compatibility between the pretraining and downstream phases.

\vspace{1mm}
\noindent
\textbf{How do SSL methods perform across domains and datasets?}  
While different SSL methods exhibit varying behaviors on in-domain datasets, their impact diminishes when transferred to cross-domain graphs. Particularly, these e-commerce graphs differ substantially from the citation network used for pretraining in both node features and structural properties, as shown in Table~\ref{tab:datasets_summary}.
Consequently, the pretrained parameters fail to provide meaningful benefits, forcing the model to relearn from scratch. These findings show that link prediction benefits from pretraining only under strong distributional alignment.

\begin{figure}[t] 
    \centering
    \includegraphics[width=0.9\columnwidth]{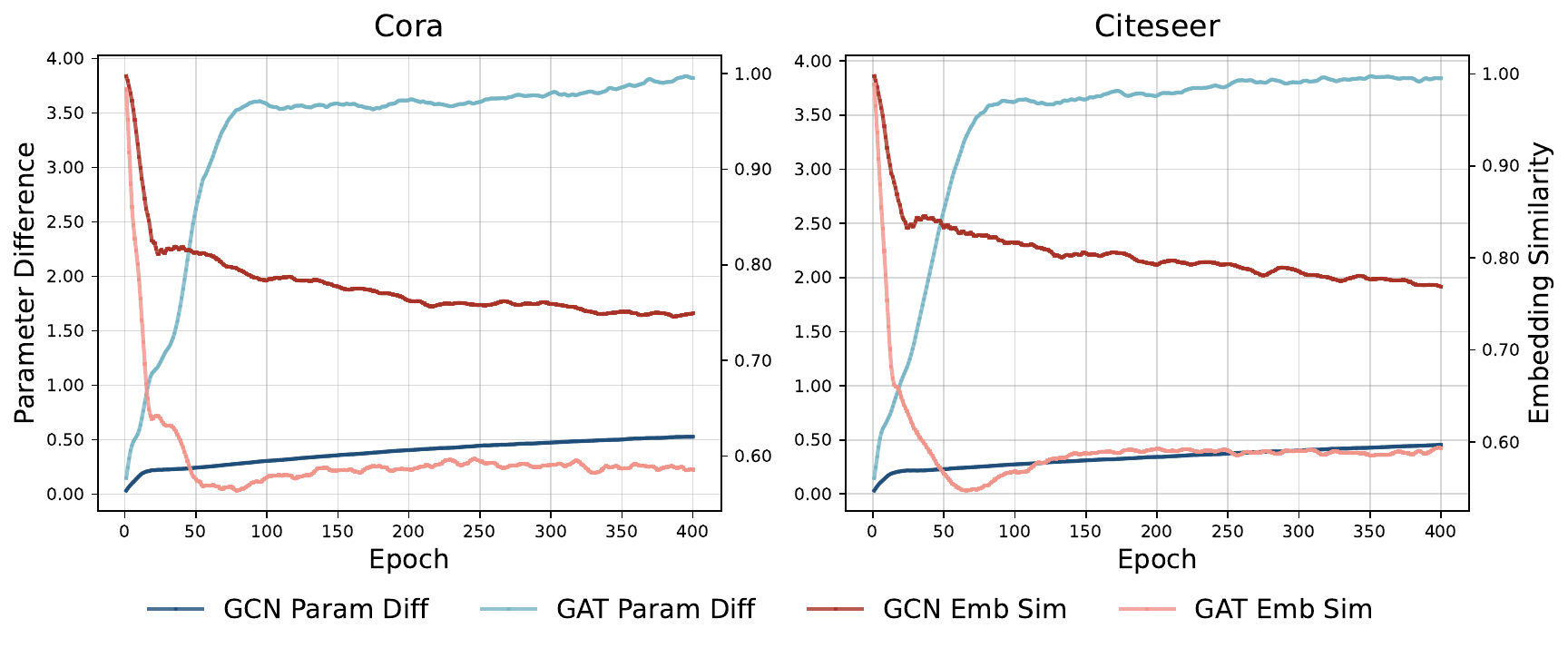} 
    \caption{GAT fine-tuning tends to overwrite pretrained parameters. X-axis denotes fine-tuning epochs. The left y-axis represents the relative difference of model parameters from the initial pretrained checkpoint. The right y-axis indicates the cosine similarity between updated node embeddings and the initial ones.}
    \label{fig:lp_curve}
    \vspace{-7mm}
\end{figure}

\vspace{1mm}
\noindent
\textbf{Does the backbone influence performance?}  
We also compare the effect of backbone architectures on transferability. Interestingly, GAT fails to benefit as much from pretraining as GCN, particularly with strong objectives like GraphMAE and LP. To understand this discrepancy, we analyze training dynamics during fine-tuning by tracking two metrics: weight drift and embedding similarity. As shown in Figure~\ref{fig:lp_curve}, GAT exhibits larger changes in model parameters and generates embeddings less similar to their pretrained counterparts, compared to GCN. This suggests that GAT adapts more aggressively during fine-tuning, potentially overwriting useful pretrained knowledge due to its flexible attention mechanism.

\vspace{1mm}
\noindent
\textbf{Key takeaways.}  
\textbf{(1)} \emph{GraphMAE} and \emph{LP} demonstrate strong performance gains on in-domain data, whereas all methods have minimal impact on cross-domain data.  
\textbf{(2)} \emph{VGAE} suffers from severe negative transfer, underscoring the importance of aligning pretraining and downstream architectures. 
\textbf{(3)} \emph{GCN} benefits significantly from pretraining, while \emph{GAT} tends to overwrite pretrained knowledge, reducing the utility of pretraining.

%% file: sections/conclusion.tex
\label{sec:conclusion}

In this work, we explored a range of self-supervised learning methods for graph pretraining, evaluating node classification and link prediction across different architectures and adaptation strategies. Our work highlights the use of LLM-derived node features, along with the unprecedented data scale present in pretraining.
Our results suggest that generative SSL methods are typically more robust in transfer settings, with the feature-reconstruction approach, GraphMAE, consistently achieving the best overall performance. We also identified several other factors influencing transfer performance, such as model architecture and adaptation methods.

Our findings provide valuable insights into future research directions in this field. First, the contrast between GraphMAE and other approaches supports its use as a stable and reliable choice when designing transferable graph models, particularly under limited computational budgets. Developing alternative methods that leverage masked feature reconstruction is also a promising direction for improving transferability.
Second, the significant differences in behavior among various SSL methods, especially between generative and contrastive approaches, suggest substantial distinctions in their learned representations. Empirically, these discrepancies highlight the potential of hybrid approaches that integrate the strengths of various methods. This is particularly crucial for foundation models, which aim to support a wide range of downstream tasks with a shared backbone. Following this, an essential research direction is to explore optimal adaptation strategies that effectively utilize pretrained knowledge, such as prompt tuning and in-context learning.
Finally, the differing behaviors of GCN and GAT in adapting to new tasks underscore the need for a deeper investigation into the underlying mechanisms driving these variations. Understanding these factors could lead to the development of refined architectures that maximize the benefits of pretraining.

A primary limitation of our work lies in the choice of language model (LM) embeddings. We employed SentenceBERT to extract node features due to its strong empirical performance and widespread use in prior graph learning and information retrieval literature. However, different encoders may yield alternative behaviors and transfer dynamics. Exploring these variations remains an important direction for future investigation.

%% file: sections/appendix.tex
\section{Detailed Experiment Settings}

\subsection{Hyperparameters}
Table~\ref{tab:ssl_hyperparams} summarizes the hyperparameters used for each SSL method.

\begin{table}[h]
\centering
\caption{Hyperparameter settings for SSL methods.}
\label{tab:ssl_hyperparams}
\scriptsize
\begin{tabular}{l l}
\toprule
\textbf{Method} & \textbf{Hyperparameters} \\
\midrule
GraphMAE &
\begin{tabular}[t]{@{}l@{}}
Node mask rate: 0.5,\quad $\alpha$: 3,\quad Decoder: GAT \\
Decoder layers: 1,\quad Dim: 384,\quad Dropout: 0.2 \\
Learning rate: $1\text{e}^{-3}$,\quad Weight decay: 0
\end{tabular} \\
\midrule
LP &
\begin{tabular}[t]{@{}l@{}}
Edge batch size: 4096,\quad Neg ratio: 1:1 \\
MLP layers: 3,\quad Dim: 384,\quad Dropout: 0.2 \\
Learning rate: $1\text{e}^{-4}$,\quad Weight decay: 0
\end{tabular} \\
\midrule
VGAE &
\begin{tabular}[t]{@{}l@{}}
Edge batch size: 4096,\quad Neg ratio: 1:1 \\
Dropout: 0.2,\quad Learning rate: $1\text{e}^{-5}$,\quad Weight decay: 0
\end{tabular} \\
\midrule
GRACE &
\begin{tabular}[t]{@{}l@{}}
Feature drop: 0.2,\quad Edge drop: 0.2,\quad Dropout: 0.2 \\
Learning rate: $1\text{e}^{-4}$,\quad Weight decay: 0
\end{tabular} \\
\midrule
DGI &
\begin{tabular}[t]{@{}l@{}}
Edge batch size: 4096,\quad Dropout: 0.2 \\
Learning rate: $1\text{e}^{-5}$,\quad Weight decay: 0
\end{tabular} \\
\bottomrule
\end{tabular}
\end{table}

For both GCN and GAT, we employ two convolutional layers with a hidden dimension of 384. Normalization techniques such as BatchNorm and LayerNorm are not applied, as we empirically find them ineffective.

\subsection{Graph In-context Learning for Few-shot Classification}
\label{sec:icl__}
We realize in-context node classification following the appraoch in~\cite{song2024pure}. Below is a brief introduction:

For an N-way K-shot classification task, the original graph \(G_{\text{ori}}\) is augmented by introducing \(N\) class nodes representing the classes in the given dataset. Each of these nodes is linked to \(K\) support examples from the training set, forming an expanded graph, denoted as \(G_{\text{aug}}\). 
The augmented graph is then fed into the pretrained GNNs, leveraging their pretrained knowledge to propagate features across both standard and class nodes.
The resulting node embeddings capture informative representations for both standard and class nodes. To classify an unseen test instance, we compute the cosine similarity between its embedding and those of the class nodes. Given the computed embeddings \( \{ s_1, s_2, \dots, s_N \} \) for the \(N\) class nodes and the embedding \( t \) for the test node, the assigned class label is determined as:

\begin{equation}
\hat{y} = \arg\max_{i \in \{1, \dots, N\}} \cos(t, s_i).
\end{equation}